\documentclass[11pt]{article}
\usepackage{acl2016}
\usepackage{amsmath}
\usepackage{amssymb}
\usepackage{times}
\usepackage{url}
\usepackage{latexsym}
\usepackage{graphicx}
\usepackage{booktabs}
\usepackage{multirow}

\aclfinalcopy % Uncomment this line for the final submission
\title{Does Multimodality Help Human and Machine for Translation and Image Captioning?}

\author{Ozan Caglayan$^{1,3}$, Walid Aransa$^1$, Yaxing Wang$^2$, Marc Masana$^2$, \\
        \bf Mercedes Garc\'ia-Mart\'inez$^1$, Fethi Bougares$^1$, Lo\"ic Barrault$^1$ and Joost van de Weijer$^2$\\
$^1$ LIUM, University of Le Mans, $^2$ CVC, Universitat Autonoma de Barcelona, \\$^3$ Galatasaray University\\
    {\tt $^1$FirstName.LastName@lium.univ-lemans.fr} \\
    {\tt $^2$\{joost,mmasana,yaxing\}@cvc.uab.es} \\
    {\tt $^3$ocaglayan@gsu.edu.tr}}

\date{}

\begin{document}
\maketitle

\begin{abstract}
This paper presents the systems developed by LIUM and CVC for the WMT16 Multimodal Machine Translation challenge.
We explored various comparative methods, namely phrase-based systems and attentional recurrent neural networks models trained using monomodal or multimodal data.
% FIXME: is this really "human machine translation" or should be human translation?
We also performed a human evaluation in order to estimate the usefulness of multimodal data for human machine translation and image description generation.
Our systems obtained the best results for both tasks according to the automatic evaluation metrics BLEU and METEOR.
\end{abstract}

\section{Introduction}
%!TEX root = wmt16_multimodal_LIUMCVC.tex

Recently, deep learning has greatly impacted the natural language processing field as well as computer vision.
Machine translation (MT) with deep neural networks (DNN), proposed by \cite{Kalchbrenner2013,Sutskever2014} and \cite{Bahdanau2014} competed successfully in the last year's WMT evaluation campaign \cite{bojar-EtAl:2015:WMT}.

In the same trend, generating descriptions from images using DNNs has been proposed by \cite{Elliott2015}.
Several attempts have been made to incorporate features from different modalities in order to help the automatic system to better model the task at hand \cite{Elliott2015,Kiros2014a,Kiros2014b}.

This paper describes the systems developed by LIUM and CVC who participated in the two proposed tasks for the WMT 2016 Multimodal Machine Translation evaluation campaign: Multimodal machine translation (Task 1) and multimodal image description (Task 2).

The remainder of this paper is structured in two parts: The first part (section~\ref{sec:multimt}) describes the architecture of the four systems (two monomodal and two multimodal) submitted for Task 1.
The standard phrase-based SMT systems based on Moses are described in section~\ref{sec:pbsmt} while the neural MT systems are described in section~\ref{sec:nmt} (monomodal) and section~\ref{sec:mnmt} (multimodal).
The second part (section~\ref{sec:multicaption}) contains the description of the two systems submitted for Task 2: The first one is a monomodal neural MT system similar to the one presented in section~\ref{sec:nmt}, and the second one is a multimodal neural machine translation (MNMT) with shared attention mechanism.

In order to evaluate the feasibility of the multimodal approach, we also asked humans to perform the two tasks of this evaluation campaign. Results show that the additional English description sentences improved performance while the straight-forward translation of the sentence without the image did not provide as good results. The results of these experiments are presented in section~\ref{sec:human}.

%----------------
\section{Multimodal Machine Translation}
\label{sec:multimt}

This task consists in translating an English sentence that describes an image into German, given the English sentence itself and the image that it describes.

\subsection{Phrase-based System}
\label{sec:pbsmt}
%!TEX root = wmt16_multimodal_LIUMCVC.tex

Our baseline system for task 1 is developed following the standard phrase-based Moses pipeline as described
in \cite{Moses:2007:acl}, SRILM \cite{Stolcke:2002qv}, KenLM \cite{Heafield-kenlm:2011}, and GIZA++ \cite{Och:2003:cl}.
This system is trained using the data provided by the organizers and tuned using MERT~\cite{Och:2003:mert} to maximize
 {\sc Bleu} \cite{Papineni:2002:acl} and {\sc Meteor} \cite{Lavie:2007:acl} scores on the validation set.

We also used Continuous Space Language Model\footnotemark\footnotetext{\texttt{github.com/hschwenk/cslm-toolkit}} (CSLM) ~\cite{schwenk10} with the auxiliary features support as proposed by~\cite{aransa15}. This CSLM architecture allows us to use sentence-level features for each line in the training data (i.e. all n-grams in the same sentence will have the same auxiliary features). By this means, better context specific LM estimations can be obtained. 

We used four additional scores to rerank 1000-best outputs of our baseline system: The first two scores are obtained from two separate CSLM(s) trained on the target side (i.e. German) of the parallel training corpus and each one of the following auxiliary features:
\begin{itemize}
 \item \textbf{VGG19-FC7 image features:} The auxiliary feature used in the first CSLM are the image features provided by the organizers which are extracted
from the FC7 layer (relu7) of the VGG-19 network \cite{simonyan2014very}. This allows us to train a multimodal CSLM that uses additional context learned from the image features.

 \item \textbf{Source side sentence representation vectors:} We used the method described in ~\cite{le2014distributed} to compute continuous space representation vector 
for each source (i.e. English) sentence that will be provided to the second CSLM as auxiliary feature. The idea behind this is to condition our target language model on the source side as additional context.
\end{itemize}

The two other scores used for n-best reranking are the log probability computed
by our NMT system that will be described in the following section and the score obtained by a Recurrent Neural Network Language Model (RNNLM) \cite{mikolov2010recurrent}. The weights of the original moses features and our additional features were optimized to maximize
the BLEU score on the validation set. 

%\begin{figure}[htbp] %  figure placement: here, top, bottom, or page
  %\centering
   %\includegraphics[width=7.7cm]{figures/nnlmaux3.pdf} 
   %\caption{CSLM architecture with auxiliary feature vectors.}
   %\label{fig:cslm_aux}
%\end{figure}

\subsection{Neural MT System}
\label{sec:nmt}
%!TEX root = wmt16_multimodal_LIUMCVC.tex

The fundamental model that we experimented\footnote{\tt github.com/nyu-dl/dl4mt-tutorial} is an attention based encoder-decoder approach \cite{Bahdanau2014} except some notable changes in the recurrent decoder called Conditional GRU.

We define by $X$ and $Y$, a source sentence of length $N$ and a target sentence of length $M$ respectively. Each source and target word is represented with an embedding vector of dimension $E_X$ and $E_Y$ respectively:
\begin{gather}
    X = (x_{1},x_{2}, ... ,x_{N}), x_{i} \in{\mathbb{R}^{E_X}} \\
    Y = (y_{1},y_{2}, ... ,y_{M}), y_{j} \in{\mathbb{R}^{E_Y}}
\end{gather}

A bidirectional recurrent encoder reads an input sequence $X$ in forwards and backwards to produce two sets of hidden states based on the current input and the previous hidden state. An annotation vector $h_{i}$ for \textit{each} position $i$ is then obtained by concatenating the produced hidden states.

%\begin{equation}
%h_{t} = \begin{bmatrix}
%	\vec{h_{t}} \\
%    \cev{h_{t}}
%    \end{bmatrix}
%\end{equation}

%TODO: We need to detail a bit more in here about the decoder
% Initial state out of the mean context vector, etc..
An attention mechanism, implemented as a simple fully-connected feed-forward neural network, accepts the hidden state $h_t$ of the decoder's recurrent layer and one input annotation at a time, to produce the attention coefficients.
A softmax activation is applied on those attention coefficients to obtain the attention weights used to generate the weighted annotation vector for time $t$.
The initial hidden state $h_0$ of the decoder is determined by a feed-forward layer receiving the mean annotation vector.

We use Gated Recurrent Unit (GRU) \cite{Chung2014} activation function for both recurrent encoders and decoders.

\subsubsection{Training}
\label{sec:task1_train}
%!TEX root = wmt16_multimodal_LIUMCVC.tex

We picked the following hyperparameters for all NMT systems both for Task 1 and Task 2. All embedding and recurrent layers have a dimensionality of 620 and 1000 respectively.
We used Adam as the stochastic optimizer with a minibatch size of 32, Xavier weight initialization \cite{glorotxavier} and L2 regularization with $\lambda=0.0001$ except the monomodal Task 1 system for which the choices were Adadelta, sampling from $\mathcal{N}(0, 0.01)$ and L2 regularization with $\lambda=0.0005$ respectively.

The performance of the network is evaluated on the validation split using BLEU after each 1000 minibatch updates and the training is stopped if BLEU does not improve for 20 evaluation periods. The training times were 16 and 26 hours respectively for monomodal and multimodal systems on a Tesla K40 GPU.

Finally, we used a classical left to right beam-search with a beam size of 12 for sentence generation during test time.

\subsection{Data}
\label{sec:task1_data}
%!TEX root = wmt16_multimodal_LIUMCVC.tex

% Data details, preprocessing, compound splitting
Phrase-based and NMT systems for Task 1 are trained using the dataset provided by the organizers and described in Table~\ref{tab:task1_data}. This dataset consists of 29K parallel sentences (direct translations of image descriptions from English to German) for training, 1014 for validation and finally 1000 for the test set. We preprocessed the dataset using the punctuation normalization, tokenization and lowercasing scripts from Moses. In order to generalize better over the compound structs in German, we trained and applied a compound splitter\footnote{\texttt{github.com/rsennrich/wmt2014-scripts}} \cite{sennrich2015joint} over the German vocabulary of training and validation sets. This reduces the target vocabulary from 18670 to 15820 unique tokens. During translation generation, the splitted compounds are stitched back together.

\begin{table}[ht]
\centering
\resizebox{0.6\columnwidth}{!}{%
  \begin{tabular}{@{}lcc@{}}
    \toprule
    Side    & Vocabulary & Words \\ \midrule
    English & 10211      & 377K  \\
    German  & 15820      & 369K
  \end{tabular}%
}
\caption{Training Data for Task 1.}
\label{tab:task1_data}
\end{table}

\subsection{Results and Analysis}
\label{sec:task1_results}
%!TEX root = wmt16_multimodal_LIUMCVC.tex

% Results for Task1

The results of our phrase-based baseline and the four submitted systems are presented in Table~\ref{tbl:task1_results}.
The \textbf{BL+4Features} system is the rescoring of the baseline 1000-best output using all the features described in \ref{sec:pbsmt} while 
\textbf{BL+3Features} is the same but excluding FC7 image features.
Overall, we were able to improve test set scores by around 0.4 and 0.8 on METEOR and BLEU respectively over a strong phrase-based baseline
using auxiliary features.
%We noticed an improvement on validation set when we used 4Features but not on test set.

Regarding the NMT systems, the monomodal NMT achieved a comparative BLEU score of 32.50 on the test set compared to
33.45 of the phrase-based baseline. The multimodal NMT system that will be described in section~\ref{sec:mnmt}, obtained
relatively lower scores when trained using Task 1's data.
% TODO: I think we need to comment more on this.

\begin{table*}[ht]
\centering
\resizebox{0.7\textwidth}{!}{%
\begin{tabular}{@{}lcccc@{}}
\toprule
\multirow{2}{*}{System Description} & \multicolumn{2}{c}{Validation Set} & \multicolumn{2}{c}{Test Set} \\ \cmidrule(l){2-5}
                                    & METEOR (norm)    & BLEU            & METEOR (norm) & BLEU         \\ \midrule
Phrase-based Baseline (BL)          & 53.71 (58.43)    & 35.61           & 52.83 (57.37) & 33.45        \\ \midrule
BL+3Features                        & 54.29 (58.99)    & 36.52           & 53.19 (57.76) & 34.31        \\
BL+4Features                        & 54.40 (59.08)    & 36.63           & 53.18 (57.76) & 34.28        \\
Monomodal NMT                       & 51.07 (54.87)    & 35.93           & 49.20 (53.10) & 32.50        \\
Multimodal NMT                      & 44.55 (47.97)    & 28.06           & 45.04 (48.52) & 27.82

\end{tabular}}
\caption{BLEU and METEOR scores on detokenized outputs of baseline and submitted Task 1 systems. The METEOR scores in parenthesis are computed with \texttt{-norm} parameter.}
\label{tbl:task1_results}
\end{table*}

%%%%% TASK 2
\section{Multimodal Image Description Generation}
\label{sec:multicaption}

The objective of Task 2 is to produce German descriptions of images given the image itself and one or more English descriptions as input.

\subsection{Visual Data Representation}
\label{sec:visual_repr}
%!TEX root = wmt16_multimodal_LIUMCVC.tex

To describe the image content we make use of Convolutional Neural Networks (CNN). In a breakthrough work, Krizhevsky et al.~\cite{krizhevsky2012imagenet} convincingly show that CNNs yield a far superior image representation compared to previously used hand-crafted image features. Based on this success an intensified research effort started to further
 improve the representations based on CNNs. The work of Simonyan and Zisserman~\cite{simonyan2014very} improved the network by breaking up large convolutional features into multiple layers of small convolutional features, which allowed to train a much deeper network. The organizers provide these features to all participants. More precisely they provide the features from the fifth convolutional layer, and the features from the second fully connected layer of VGG-19. Recently, Residual Networks (ResNet) have been proposed~\cite{he2015deep}. These networks learn residual functions which are constructed by adding skip layers (or projection layers) to the network. These skip layers prevent the vanishing gradient problem, and allow for much deeper networks (over hundred layers) to be trained.
 
\begin{figure}
\centering
\includegraphics[scale=0.33]{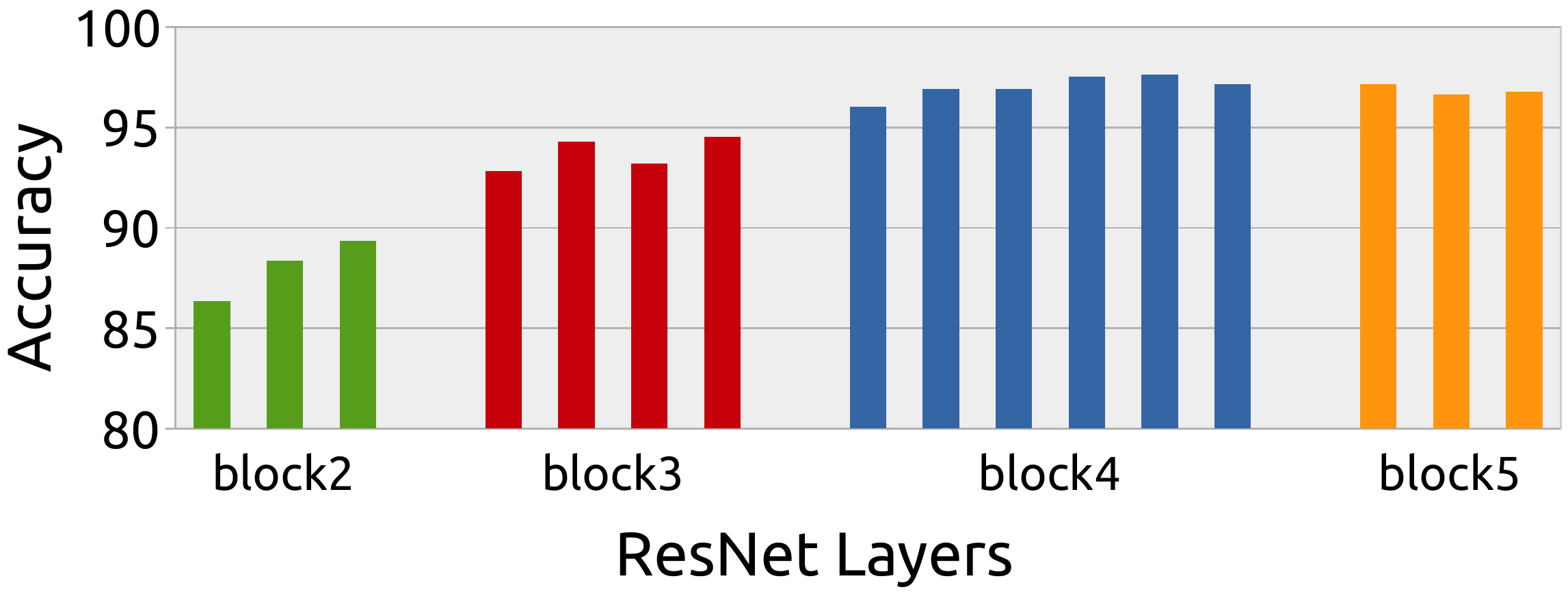}
  \caption{Classification accuracy on a subset of SUN scenes~\cite{xiao2010sun} for ResNet-50: The colored groups represent the building blocks while the bars inside are the stacked blocks~\cite{he2015deep}.}
\label{fig:resnet50_study}
\end{figure}

To select the optimal layer for image representation we performed an image classification task on a subsection of images from SUN scenes~\cite{xiao2010sun}. We extract the features from the various layers of ResNet-50 and evaluate the classification performance (Figure \ref{fig:resnet50_study}). The results increase during the first layers but stabilize from Block-4 on. Based on these results and considering that a higher spatial resolution is better, we have selected layer 'res4f\_relu' (end of Block-4, after ReLU) for the experiments on multimodal MT.
We also compared the features from different networks on the task of image description generation with the system of Xu et al.~\cite{xu2015show}. The results for generating English descriptions (Table~\ref{tab:visual_feats}) show a clear performance improvement from VGG-19 to ResNet-50, but comparable results are obtained when going to ResNet-152. Therefore, given the increase in computational cost, we have decided to use ResNet-50 features for our submission.
% one paragraph on the results. The choice of the layer and the TABLE of Xu

\begin{table}[t]
\centering
\resizebox{0.48\textwidth}{!}{%
\begin{tabular}{rcccc}
\hline
Network		& BLEU-1 & BLEU-2 & BLEU-3 & BLEU-4 \\
\hline
VGG-19		& 58.2 & 31.4 & 18.5 & 11.3  \\
ResNet-50	& 68.4 & 45.2 & 30.9 & 21.1  \\
ResNet-152	& 68.3 & 44.9 & 30.7 & 21.1  \\
\hline
\end{tabular}}
\caption{BLEU scores for various deep features on the image description generation task using the system of Xu et al.~\cite{xu2015show}.}
\label{tab:visual_feats}
\end{table}

\subsection{Multimodal NMT System}
\label{sec:mnmt}
%!TEX root = wmt16_multimodal_LIUMCVC.tex

\newcommand{\ig}{i}
\newcommand{\ug}{z}
\newcommand{\rg}{r}
\newcommand{\hg}{h}

\newcommand{\mat}[1]{#1}
\newcommand{\scu}[1]{\sc{\underline{#1}}}
\newcommand{\vect}[1]{#1}
\newcommand{\uvect}[1]{\underline{#1}}

\newcommand{\E}{E}
\newcommand{\U}{U}
\newcommand{\W}{W}

The multimodal NMT system is an extension of \cite{xu2015show} and the monomodal NMT system described in Section~\ref{sec:nmt}.

The model involves two GRU layers and an attention mechanism.
The first GRU layer computes an intermediate representation $\vect{s}^{'}_j$ as follows:
\begin{eqnarray}
  \vect{s}^{'}_j =& (1 - \ug^{'}_j) \odot \underline{\vect{s}}^{'}_j + \ug^{'}_j \odot \vect{s}_{j-1} \\
  \underline{\vect{s}}^{'}_j =& \tanh(\W^{'}\E[y_{j-1}]+\rg^{'}_j \odot (\U^{'}\vect{s}_{j-1})) \\
  \rg^{'}_j =& \sigma(\W^{'}_r \E[y_{j-1}] + \U^{'}_r\vect{s}_{j-1}) \\
  \ug^{'}_j =& \sigma(\W^{'}_z \E[y_{j-1}] + \U^{'}_z\vect{s}_{j-1})
\end{eqnarray}
where $\E$ is the target word embedding, $\underline{\vect{s}}^{'}_j$ is the hidden state, $\rg^{'}_j$ and $\ug^{'}_j$ are the reset and update gate activations. $\W^{'}$, $\U^{'}_r$, $\W^{'}_r$, $\U^{'}_r$, $\W^{'}_z$ and $\U^{'}_z$ are the parameters to be learned.

A shared attention layer similar to \cite{firat2016multi} that consists of a fully-connected feed-forward network is used to compute a set of modality specific attention coefficients $e^{mod}_{ij}$ at each timestep $j$:
\begin{equation}
	%%\vect{e}_{ij} = \vect{v}_a^\intercal \tanh \left( \U_a\vect{s}_j + \W_a \vect{h}_i \right)
  \vect{e}^{mod}_{ij} = \vect{U}_{att} \tanh(\W_{catt} \vect{h}^{mod}_i + \W_{att}\vect{s}^{'}_j)
\end{equation}
The attention weight between source modality context $i$ and target word $j$ is computed by applying a softmax on
$\vect{e}^{mod}_{ij}$ :
\begin{eqnarray}
  \alpha_{ij} = \frac {\exp(e^{txt}_{ij})} {\sum_{k=1}^{N} \exp(e^{txt}_{kj})}\\
  \beta_{ij} = \frac {\exp(e^{img}_{ij})} {\sum_{k=1}^{196} \exp(e^{img}_{kj})}
\end{eqnarray}
The final multimodal context vector $\vect{c}_j$ is obtained as follows:
\begin{equation}
  \vect{c}_j = \tanh (\sum_{i=1}^{N}  \alpha_{ij} \, \vect{h}^{txt}_i + \sum_{i=1}^{196} \beta_{ij} \, \vect{h}^{img}_i)
\end{equation}

The second GRU generates $\vect{s}_j$ from the intermediate representation $\vect{s}^{'}_j$ and the context vector $\vect{c}_j$ as follows:
\begin{eqnarray}
	\vect{s}_j &=& (1 - \ug_j) \odot \uvect{s}_j + \ug_j \odot \vect{s}^{'}_j \\
	\uvect{s}_j &=& \tanh ( \W \vect{c}_j + \rg_j \odot (\U\vect{s}^{'}_j) ) \\
	\rg_j &=& \sigma(\W_r \vect{c}_j + \U_r\vect{s}^{'}_j) \\
	\ug_j &=& \sigma(\W_z \vect{c}_j + \U_z\vect{s}^{'}_j)
\end{eqnarray}
where $\underline{\vect{s}}^{'}_j$ is the hidden state, $\rg_j$ and $\ug_j$ are the reset and update gate activations. $\W$, $\U_r$, $\W_r$, $\U_r$, $\W_z$ and $\U_z$ are the parameters to be learned.

Finally, in order to compute the target word, the following formulations are applied:
\begin{eqnarray}
\vect{o}_j = \mat{L}_o \tanh (E[y_{j-1}] + \mat{L}_{s}\vect{s}_j + \mat{L}_{c}\vect{c}_j)\\
P(y_j| y_{j-1}, \vect{s}_j, \vect{c}_j) = \text{Softmax}(\vect{o}_j)
\end{eqnarray}
where $\mat{L}_o$, $\mat{L}_{s}$ and $\mat{L}_{c}$ are trained parameters.

\begin{figure*}[hbtp!]
\centering
  \includegraphics[width=\textwidth]{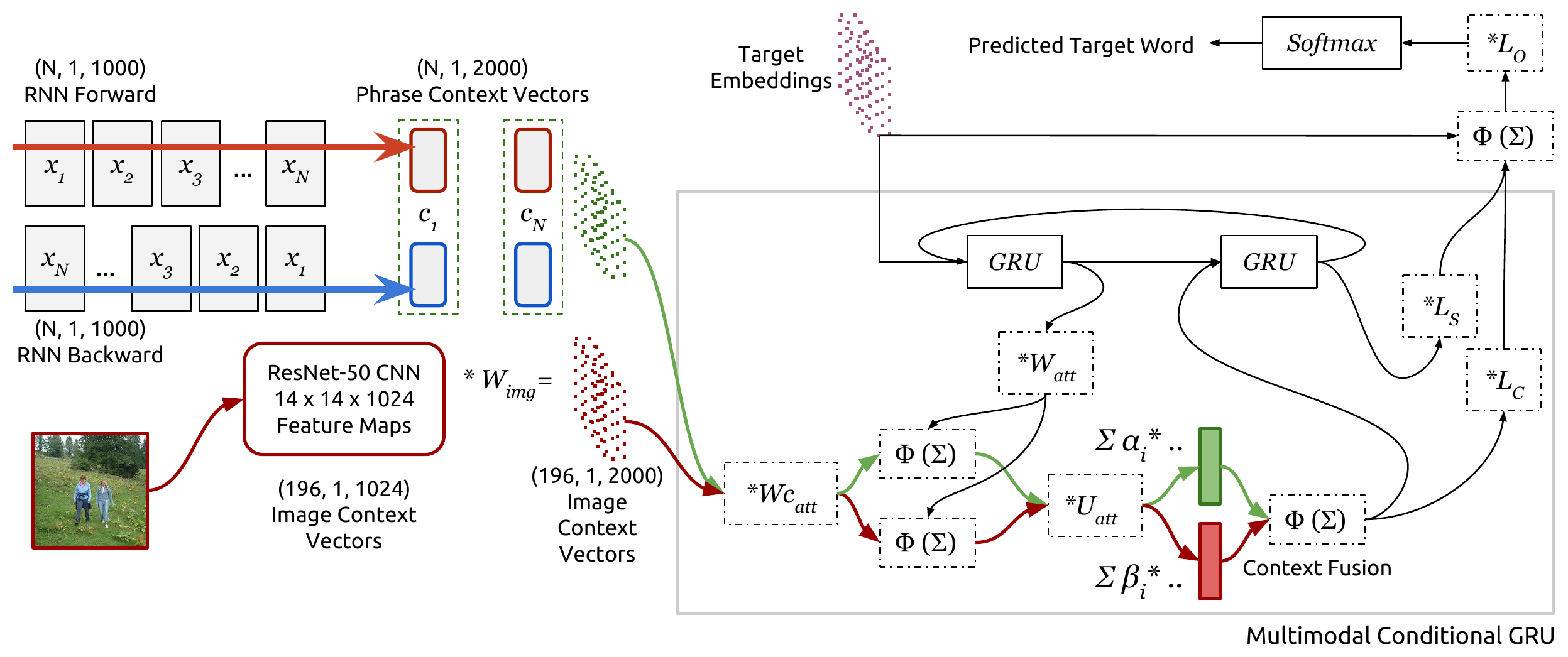}
  \caption{The architecture of the multimodal NMT system. The boxes with $*$ refers to a linear transformation while $\Phi(\Sigma)$ means a $tanh$ applied over the sum of the inputs. The figure depicts
  a running instance of the network over a single example.}
  \label{fig:mnmt_arch}
\end{figure*}

% if those equation take too much place, we will put them on the additional material.

\subsubsection{Generation}
\label{sec:task2_train}
Since we are provided 5 source descriptions for each image in order to generate a single German description, we let the NMT generate a German description for each source and pick the one with the highest probability and preferably without an UNK token.

\subsection{Data}
\label{sec:task2_data}
%!TEX root = wmt16_multimodal_LIUMCVC.tex
% Data details, preprocessing
The organizers provided an extended version of the Flickr30K Entities dataset \cite{Elliott2016} which contains 5 \emph{independently} crowd-sourced German descriptions for each image in addition to the 5 English descriptions originally found in the dataset. It is possible to use this dataset either by considering the cross product of 5 source and 5 target descriptions (a total of 25 description pairs for each image) or by only taking the 5 pairwise descriptions leading to 725K and 145K training pairs respectively. We decided to use the smaller subset of 145K sentences.

\begin{table}[ht]
\centering
\resizebox{0.6\columnwidth}{!}{%
\begin{tabular}{@{}lcc@{}}
\toprule
Side    & Vocabulary & Words \\ \midrule
English & 16802      & 1.5M  \\
German  & 10000      & 1.3M 
\end{tabular}}
\caption{Training Data for Task 2.}
\label{tbl:task2_data}
\end{table}

The preprocessing is exactly the same as Task 1 except that we only kept sentence pairs with sentence lengths $\in{[3,50]}$ and with a ratio of at most 3. This results in a final training dataset of 131K sentences (Table \ref{tbl:task2_data}). We picked the most frequent 10K German words and replaced the rest with an UNK token for the target side. Note that compound splitting was not done for this task.

\subsection{Results and Analysis}
\label{sec:task2_results}
%!TEX root = wmt16_multimodal_LIUMCVC.tex

\begin{table}[ht]
\centering
\resizebox{0.47\textwidth}{!}{%
\begin{tabular}{@{}rcccc@{}}
\toprule
\multirow{2}{*}{System}         & \multicolumn{2}{c}{Validation} & \multicolumn{2}{c}{Test} \\ \cmidrule(l){2-5} 
                                & METEOR            & BLEU           & METEOR         & BLEU        \\ \midrule
\multicolumn{1}{r}{Monomodal}   & 36.3              & 24.0           & 35.1           & 23.8        \\
\multicolumn{1}{r}{Multimodal}  & 34.4              & 19.3           & 32.3           & 19.2        \\ \bottomrule
\end{tabular}}
\caption{BLEU and METEOR scores of our NMT based submissions for Task 2.}
\label{tbl:task2_results}
\end{table}
% TODO: We can also put VGG results.
As we can see in Table~\ref{tbl:task2_results}, the multimodal system does not surpass monomodal NMT system.
Several explanations can clarify this behavior. First, the architecture is not well suited for integrating image and text representations.
This is possible as we did not explore all the possibilities to benefit from both modalities.
Another explanation is that the image context contain too much irrelevant information which cannot be discriminated by the lone attention mechanism.
This would need a deeper analysis of the attention weights in order to be answered.

%----------------
% Can be moved somewhere else depending on the content
\section{Human multimodal description}
\label{sec:human}
%!TEX root = wmt16_multimodal_LIUMCVC.tex

To evaluate the importance of the different modalities for the image description generation task, we have performed an experiment where we replace the computer algorithm with human participants. The two modalities are the five English description sentences, and the image. The output is a single description sentence in German. The experiment asks the participants for the following tasks:
\begin{itemize}
\item Given both the image and the English descriptions: '\textit{Describe the image in one sentence in German. You can get help from the English sentences provided.}'
\vspace{-0.5em}
\item Given only the image: '\textit{Describe the image in one sentence in German.}'
\vspace{-0.5em}
\item Given only one English sentence: '\textit{Translate the English sentence into German.}'
\end{itemize}
The experiment was performed by 16 native German speakers proficient in English with age ranging from 23 to 54 (coming from Austria, Germany and Switzerland, of which 10 are female and 6 male). The experiment is performed on the first 80 sentences of the validation set. Participants performed 10 repetitions for each task, and not repeating the same image across tasks. The results of the experiments are presented in Table~\ref{tab:human_exp}. For humans, the English description sentences help to obtain better performance. Removing the image altogether and providing only a single English description sentence results in a significant drop. We were surprised to observe such a drop, whereas we expected good translations to obtain competitive results. In addition, we have provided the results of our submission on the same subset of images; humans clearly obtain better performance using METEOR metrics, but our approach is clearly outperforming on the BLEU metrics. The participants were not trained on the train set before performing the tasks, which could be one of the reasons for the difference. Furthermore, given the lower performance of only translating one of the English description sentences on both metrics, it could possibly be caused by existing biases in the data set.

\begin{table}[t]
\centering
\resizebox{0.48\textwidth}{!}{%
\begin{tabular}{rccccc}
\hline
Method				& BLEU-1 & BLEU-2 & BLEU-3 & BLEU-4 & METEOR\\
\hline
Image + sentences	& 54.30 & 35.95 & 23.28 & 15.06 & 39.16 \\
Image only			& 51.26 & 34.74 & 22.63 & 15.01 & 38.06 \\
Sentence only		& 39.37 & 23.27 & 13.73 &  8.40 & 32.98 \\
\hline
Our system			& 60.61 & 44.35 & 31.65 & 21.95 & 33.59 \\
\hline
\end{tabular}}
\caption{BLEU and METEOR scores for human description generation experiments.}
\label{tab:human_exp}
\end{table}

%----------------
\section{Conclusion and Discussion}
%!TEX root = wmt16_multimodal_LIUMCVC.tex

We have presented the systems developed by LIUM and CVC for the WMT16 Multimodal Machine Translation challenge.
Results show that integrating image features into a multimodal neural MT system with shared attention mechanism does not yet surpass the performance obtained with a monomodal system using only text input. However, our multimodal systems do improve upon an image captioning system (which was expected). 
The phrase-based system can benefit from rescoring with multimodal neural language model as well as rescoring with a neural MT system.

We have also presented the results of a human evaluation performing the same tasks as proposed in the challenge.
The results are rather clear: image captioning can benefit from multimodality.

\section*{Acknowledgments}

This work was supported by the Chist-ERA project M2CR\footnote{\tt m2cr.univ-lemans.fr}.
We kindly thank KyungHyun Cho and Orhan Firat for providing the DL4MT tutorial as open source and Kelvin Xu for the arctic-captions\footnote{\tt github.com/kelvinxu/arctic-captions} system.
% include your own bib file like this:
\bibliography{wmt16}
\bibliographystyle{acl2016}

\end{document}